# Improving Polish to English Neural Machine Translation with Transfer Learning: Effects of Data Volume and Language Similarity

**Juuso Eronen**
Prefectural University of Kumamoto
`eronenj@pu-kumamoto.ac.jp`

**Michal Ptaszynski**
Kitami Institute of Technology
`michal@mail.kitami-it.ac.jp`

**Karol Nowakowski**
Tohoku University of
Community Service and Science
`karol@koeki-u.ac.jp`

**Zheng Lin Chia**
Kitami Institute of Technology
`chiazhenglin@gmail.com`

**Fumito Masui**
Kitami Institute of Technology
`f-masui@mail.kitami-it.ac.jp`

## Abstract

This paper investigates the impact of data volume and the use of similar languages on transfer learning in a machine translation task. We find out that having more data generally leads to better performance, as it allows the model to learn more patterns and generalizations from the data. However, related languages can also be particularly effective when there is limited data available for a specific language pair, as the model can leverage the similarities between the languages to improve performance. To demonstrate, we fine-tune mBART model for a Polish-English translation task using the OPUS-100 dataset. We evaluate the performance of the model under various transfer learning configurations, including different transfer source languages and different shot levels for Polish, and report the results. Our experiments show that a combination of related languages and larger amounts of data outperforms the model trained on related languages or larger amounts of data alone. Additionally, we show the importance of related languages in zero-shot and few-shot configurations.

## 1 Introduction

Machine translation is a vital technology that facilitates communication between people who speak different languages. However, machine translation is a challenging task that requires large amounts of high-quality data for training. Unfortunately, obtaining sufficient data for every language pair can be a difficult and expensive task (Engelson and Dagan, 1996), (Dandapat et al., 2009). Therefore, researchers have turned to transfer learning as a means of improving machine translation performance (Dabre et al., 2020).

The idea of transfer learning is to leverage the knowledge learned from one language pair to improve the performance of a model on another language pair. In most cases, transfer learning is performed from any language with a lot of available data, or by using data from related languages.

Recent research has shown that transfer learning can be an effective approach for improving machine translation performance. It is common to opt to using more data from high-resource languages for a better performance (Zoph et al., 2016). In a study by Kocmi and Bojar (Kocmi and Bojar, 2018), the authors found that the size of the transfer source dataset is more important than the relatedness of the languages. Another way to increase performance with more data is to use multiple source languages for the transfer learning (Maimaiti et al., 2019).

In addition to data quantity, relatedness between languages is also an important factor in transfer learning for machine translation. Related lan-guages share common features, such as grammatical structures and vocabulary, which can be exploited in transfer learning (Nooralahzadeh et al., 2020). In particular, related languages can be effective when there is limited data available for a specific language pair (Cotterell and Heigold, 2017).

In a study by Nguyen and Chiang (Nguyen and Chiang, 2017), the authors investigated the effectiveness of transfer learning using pre-trained models on different language pairs. They found that transfer learning was effective in improving the performance of machine translation models on related language pairs. Similarly, Dabre et al. (Dabre et al., 2017) studied the effectiveness of transfer learning for low-resource languages and found that transfer source languages falling in the same or linguistically similar language family perform the best.

In this paper, we explore the impact of data volume and the use of similar languages in a machine translation task. In practice, we fine-tune the multilingual BART (Tang et al., 2020) model for a Polish to English translation task using the OPUS-100 (Zhang et al., 2020) dataset. We evaluate the performance of the model under different transfer learning configurations, including zero-shot and few-shot configurations. Our study finds that both more data and related languages can be important for transfer learning in machine translation. Having more data can generally lead to better performance, but related languages can be particularly effective when there is limited data available for a specific language pair. Overall, this study contributes to our understanding of the importance of data quantity and language relatedness in transfer learning for machine translation.

## 2 Previous Research

### 2.1 Data Volume

Data volume is an important factor in transfer learning for machine translation. Generally, having more data available can lead to better performance. This is because more data provides the model with a larger and more diverse set of examples to learn from, which can lead to improved generalization and better performance on unseen data.

Several studies have shown the effectiveness of increased data quantity for transfer learning in machine translation. For example, in a study by Zoph et al. (Zoph et al., 2016), the authors investigated the effectiveness of using large amounts of data from high-resource languages to improve the performance of machine translation models on low-resource languages. They found that using large amounts of data from high-resource languages can lead to significant improvements in performance on low-resource languages.

Similarly, in a study by Koehn and Knowles (Koehn and Knowles, 2017), the authors investigated the effectiveness of using more data for transfer learning in machine translation across multiple language pairs. They found that using larger amounts of data generally leads to better performance, but the effectiveness of additional data decreases as the amount of data increases.

According to Kocmi and Bojar (Kocmi and Bojar, 2018), the sheer the size of the used source corpus can be more important than the relatedness of the source and target languages. They found out that Czech and Estonian sometimes worked better as a language pair than Finnish and Estonian even though the languages are not related.

Also, it has been shown that using multiple languages as the transfer source can lead to higher performance (McDonald et al., 2011). For example, both Maimati et al. (Maimaiti et al., 2019) and Chen et al. (Chen et al., 2019) showed that multi-source cross-lingual transfer can be very effective for machine translation.

### 2.2 Similar Languages

Relatedness between languages plays an important role in transfer learning for machine translation. Languages that are related or belong to the same language family often share similar grammatical structures, vocabulary, and syntax. This shared linguistic background can be exploited to improve the performance of machine translation systems (Nooralahzadeh et al., 2020).

For example, in a study by Cotterell and Heigold (Cotterell and Heigold, 2017), the authors investigated cross-lingual transfer learning for low-resource languages. They found that related languages, such as Spanish and Portuguese or Czech and Slovak, improved the performance of machine translation models compared to unrelated language pairs.

Relatedness between languages has been found to be an important factor in transfer learning for machine translation. Nguyen and Chiang (Nguyen and Chiang, 2017) found that transfer learning was particularly effective in improving the performance of machine translation models on related language pairs. This is because related languages tend to share common linguistic features, such as grammatical structures and vocabulary, which can be exploited in transfer learning.

Similarly, Dabre et al. (Dabre et al., 2017) found that transfer source languages falling in the same or linguistically similar language family perform the best for low-resource languages. This is because transfer learning can leverage the knowledge learned from the languages to improve the translation quality of the transfer target language.

Relatedness between languages has also been studied in the context of zero-shot and few-shot machine translation, where the goal is to translate between language pairs for which no or very little parallel data is available. Nooralahzadeh et al. (Nooralahzadeh et al., 2020) showed that related languages tend to perform better in zero-shot translation, where the system is trained on a transfer source language and tested on a transfer target language with no parallel data between them.

Relatedness between languages is also important when there is limited data available for a specific language pair. Transfer learning can be particularly effective when there is a lack of parallel data, which is often the case for low-resource languages (Gaikwad et al., 2021). By using related languages, it is possible to leverage existing data and transfer knowledge across languages to improve the performance of machine translation models (Martínez-García et al., 2021).

Additionally, it has been shown that there is a correlation between the similarity of the used language pair and cross-lingual transfer efficiencyfor multiple natural language processing tasks (Lauscher et al., 2020), (Eronen et al., 2023).

## 3 Methods

In this section, we describe the methodology we used to study the impact of data volume and language similarity on transfer learning in machine translation.

We fine-tuned the multilingual BART (mBART) (Tang et al., 2020) model for the Polish-English translation task. mBART is a pre-trained language model developed by Facebook AI Research (FAIR) that is designed to improve machine translation and other sequence-to-sequence tasks across multiple languages. It is based on the BERT architecture and is trained on a diverse set of languages. mBART has achieved state-of-the-art performance on various machine translation benchmarks and has shown promising results in cross-lingual transfer learning tasks.

The fine-tuning is done using the OPUS-100 corpus. It is a large-scale parallel corpus consisting of more than 100 million sentences in over 100 languages (Zhang et al., 2020). The corpus is designed to facilitate research on multilingual natural language processing, including machine translation, cross-lingual information retrieval, and language modeling. The data is collected from various sources, including web pages, books, and subtitles, and the text is aligned at the sentence level to create parallel corpora for each language pair. Being one of the largest open parallel corpora available, the Opus-100 corpus has become a widely used benchmark dataset for multilingual machine translation and has been used in a number of studies exploring various approaches to multilingual natural language processing.

To evaluate the impact of data volume and the use of related languages, we propose five different models. First, we use a baseline model fine-tuned only on Polish. The other four models are trained in the same manner as Zoph et al.(Zoph et al., 2016) in a parent-child configuration. We fine-tune a parent model first in other languages in a translation task to English. We swap the training corpus and fine-tuning is then continued on these models on the Polish to English task.

The composition of the parent models varies in terms of language similarity, with the first parent model using Czech, a West Slavic language similar to Polish. The second model is fine-tuned in Russian, which is an East Slavic language, a slightly more distant cousin to Polish and Czech. The third model is a Slavic parent model that includes both Czech and Russian, while the fourth model is fine-tuned in German, which is not related to Polish.

To fine-tune the models on the Polish-English task, we use five different configurations. The configurations use different amounts of Polish samples, specifically zero, ten, one hundred, one thousand and ten thousand. Using zero samples means that we evaluate the models in a zero-shot configuration, in which case no Polish data is used for the fine-tuning. By using these different configurations and parent models, we can evaluate the impact of language similarity and data volume on transfer learning in machine translation.

The performance of the machine translation models was evaluated using the BLEU and METEOR metrics. BLEU (Bilingual Evaluation Understudy) is a widely used automatic evaluation metric for machine translation that measures the similarity between the machine-generated translations and the human reference translations (Papineni et al., 2002). The metric ranges from 0 to 1, with higher values indicating better translation quality. The scores are calculated based on the n-gram overlap between the machine-generated and reference translations, as well as the brevity penalty that penalizes the model for generating shorter translations than the reference translations.

METEOR (Metric for Evaluation of Translation with Explicit ORdering) is a widely used evaluation metric in machine translation research, along with BLEU (Banerjee and Lavie, 2005). METEOR is based on a combination of precision, recall, and alignment between the candidate translation and the reference translation, and also considers the fluency and adequacy of the translation. METEOR has been shown to correlate well with human judgments of translation quality and is considered a useful metric for evaluating machine translation performance.

The models were fine-tuned by using PyTorch and the Huggingface Transformers library (Wolf et al., 2020). The hardware used was an Nvidia RTX 3090 GPU.

## 4 Results and Discussion

We fine-tuned mBART in the configurations introduced earlier. The parent models were fine-tuned using one hundred thousand samples with each of the languages. These models were then additionally fine-tuned with ten thousand, one thousand, one hundred and ten samples of Polish before the evaluation step. The model evaluation scores for all configurations are presented in Table 1.

The table presents the results of the Polish-English translation experiment using different transfer languages at various shot levels of Polish. The experiment was evaluated using the two previously introduced evaluation metrics, BLEU and METEOR. The shot levels represent the amount of Polish data available for fine-tuning the model. The shot levels range from 0 shot (no Polish data) to 10k shot (10,000 samples of Polish used for fine-tuning).

It can be seen from the results that adding higher amounts of transfer target language data (Polish) clearly yield a higher performance. Having more data generally leads to better performance as larger datasets enable the model to learn more patterns and generalizations from the data, which can improve the model's ability to translate accurately. In contradiction to our expectations though, using more transfer source language data does not seem to have so much of an impact. The Slavic model is fine-tuned with twice the amount of data compared to other models as it uses both Czech and Russian data. Despite this, the performance is lower than using only Czech on zero-shot and few-shot cases and lower than using only Russian on more high-resource cases. We need to investigate the use of multiple transfer languages more in the future.

Also, the the results show that related languages are important in zero-shot and few-shot settings, where limited data is available for a given language pair. This has important implications for the development of machine translation models in low-resource scenarios, where transfer learning can be particularly effective. This is because related languages share common features, such as grammatical structures and vocabulary, which can be exploited in transfer learning to improve performance.

This effect seems to diminish however as the amount of transfer target language data (Polish) increases. In more high-resource cases, it does not seem to matter which language is used as the transfer source as with ten thousand samples of Polish, both Russian and German outperform Czech slightly despite Czech being more closely related to Polish than the other languages used. It seems like that with enough samples from the transfer target language, the model can achieve a noticeably higher scores when transferring from any language. Additionally, when the transfer source language is of high similarity (Czech) with source language (Polish), its possible to have completely zero-shot results on comparable or even higher level than in a few-shot configuration with less similar languages (Russian, Slavic).

Our results have implications for the development of machine translation models, particularly for low-resource languages. In such scenarios, related languages may be useful in improving the performance of machine translation models. Fur-thermore, our findings suggest that efforts to increase the amount of training data available for a given language pair can also lead to improved performance.

Table 1: BLEU and METEOR scores for Polish-English translation

| Source lang: Polish | 0 shot | | 10 shot | | 100 shot | | 1k shot | | 10k shot | |
|---|---|---|---|---|---|---|---|---|---|---|
| Transfer lang: | BLEU | METEOR | BLEU | METEOR | BLEU | METEOR | BLEU | METEOR | BLEU | METEOR |
| N/A | – | – | 0.45 | 0.05 | 0.01 | 0.01 | 10.43 | 0.33 | 15.42 | 0.36 |
| Czech | 11.61 | 0.35 | 14.3 | 0.41 | 13.41 | 0.37 | 14.35 | 0.42 | 17.17 | 0.41 |
| Russian | 0.42 | 0.11 | 3.16 | 0.26 | 4.86 | 0.31 | 16.44 | 0.41 | 19.42 | 0.44 |
| Slavic | 8.33 | 0.27 | 11.94 | 0.36 | 10.87 | 0.35 | 16.44 | 0.41 | 18.18 | 0.43 |
| German | 0.12 | 0.05 | 0.56 | 0.07 | 3.72 | 0.29 | 16.82 | 0.42 | 19.35 | 0.44 |

One of the main limitations of this study is that we only used one dataset and one language pair, which may limit the generalizability of our findings. The OPUS-100 dataset contains a large amount of data from many languages, but it is still a single dataset and does not fully represent the full range of available content. Similarly, while our study focused on the Polish-English language pair, it is possible that the effectiveness of transfer learning varies across other language pairs.

In the future we are planning to confirm the results with other datasets and other language pairs than Polish-English. We will also investigate the use of related languages in other NLP tasks beyond machine translation, and explore the optimal combination of relatedness and data volume in transfer learning.

Our study suggests that transfer learning can be an effective approach for improving machine translation performance, particularly in low-resource settings. However, further research is needed to investigate the generalizability of our findings to other language pairs and datasets, as well as to explore the effectiveness of transfer learning in more complex real-world settings.

## 5 Conclusions

In conclusion, our study showed that the volume of the transfer target language data and language similarity can have a significant impact on transfer learning in machine translation. Contrary to our expectations, using additional transfer source language data did not seem to make a difference. The results indicate that having more data generally leads to better performance, but related languages can be particularly effective when there is limited data available for a specific language pair. Our experiments with different parent models and fine-tuning configurations demonstrate that incorporating language similarity in transfer learning can help improve machine translation performance, especially in low-resource scenarios.

Based on our results, we recommend that researchers and practitioners consider language similarity when designing transfer learning approaches for machine translation. When there is limited data available for a specific language pair, incorporating related languages in the training data can improve performance.

In the future, we need to confirm the results also with other datasets and other language pairs. We need to investigate the use of related languages in other NLP tasks beyond machine translation, and explore the the use of multiple transfer source languages more in the future.

Overall, our study contributes to a better understanding of the factors that influence transfer learning in machine translation and provides insights into how to design effective transfer learning approaches for this task. We hope that our findings will be useful for researchers and practitioners working in the field of natural language processing and machine translation.

## References

Banerjee, Satanjeev and Alon Lavie. 2005. METEOR: An automatic metric for MT evaluation with improved correlation with human judgments. In *Proceedings of the ACL Workshop on Intrinsic and Extrinsic Evaluation Measures for Machine Translation and/or Summarization*, pages 65–72, Ann Arbor, Michigan, June. Association for Computational Linguistics.

Chen, Xilun, Ahmed Hassan Awadallah, Hany Hassan, Wei Wang, and Claire Cardie. 2019. Multi-source cross-lingual model transfer: Learning what to share. In *Proceedings of the 57th Annual Meeting of the Association for Computational Linguistics*, pages 3098–3112, Florence, Italy, July. Association for Computational Linguistics.

Cotterell, Ryan and Georg Heigold. 2017. Cross-

lingual character-level neural morphological tagging. In *Proceedings of the 2017 Conference on Empirical Methods in Natural Language Processing*, pages 748–759, Copenhagen, Denmark, September. Association for Computational Linguistics.

Dabre, Raj, Tetsuji Nakagawa, and Hideto Kazawa. 2017. An empirical study of language relatedness for transfer learning in neural machine translation. In *Pacific Asia Conference on Language, Information and Computation*.

Dabre, Raj, Chenhui Chu, and Anoop Kunchukuttan. 2020. A survey of multilingual neural machine translation. *ACM Comput. Surv.*, 53(5), September.

Dandapat, Sandipan, Priyanka Biswas, Monojit Choudhury, and Kalika Bali. 2009. Complex linguistic annotation–no easy way out! a case from bangla and hindi pos labeling tasks. In *Proceedings of the Third Linguistic Annotation Workshop (LAW III)*, pages 10–18.

Engelson, Sean P. and Ido Dagan. 1996. Minimizing manual annotation cost in supervised training from corpora. In *Proceedings of the 34th Annual Meeting on Association for Computational Linguistics*, ACL '96, page 319–326, USA. Association for Computational Linguistics.

Eronen, Juuso, Michal Ptaszynski, and Fumito Masui. 2023. Zero-shot cross-lingual transfer language selection using linguistic similarity. *Information Processing & Management*, 60(3):103250.

Gaikwad, Saurabh, Tharindu Ranasinghe, Marcos Zampieri, and Christopher M. Homan. 2021. Cross-lingual offensive language identification for low resource languages: The case of marathi.

Kocmi, Tom and Ondřej Bojar. 2018. Trivial transfer learning for low-resource neural machine translation. In *Proceedings of the Third Conference on Machine Translation: Research Papers*, pages 244–252, Brussels, Belgium, October. Association for Computational Linguistics.

Koehn, Philipp and Rebecca Knowles. 2017. Six challenges for neural machine translation. In *Proceedings of the First Workshop on Neural Machine Translation*, pages 28–39, Vancouver, August. Association for Computational Linguistics.

Lauscher, Anne, Vinit Ravishankar, Ivan Vulić, and Goran Glavaš. 2020. From zero to hero: On the limitations of zero-shot language transfer with multilingual Transformers. In *Proceedings of the 2020 Conference on Empirical Methods in Natural Language Processing (EMNLP)*, pages 4483–4499, Online, November. Association for Computational Linguistics.

Maimaiti, M., Yang Liu, Huanbo Luan, and Maosong Sun. 2019. Multi-round transfer learning for low-resource nmt using multiple high-resource languages. *ACM Transactions on Asian and Low-Resource Language Information Processing (TALLIP)*, 18:1 – 26.

Martínez-García, Antonio, Toni Badia, and Jeremy Barnes. 2021. Evaluating morphological typology in zero-shot cross-lingual transfer. In *Proceedings of the 59th Annual Meeting of the Association for Computational Linguistics and the 11th International Joint Conference on Natural Language Processing (Volume 1: Long Papers)*, pages 3136–3153, Online, August. Association for Computational Linguistics.

McDonald, Ryan, Slav Petrov, and Keith Hall. 2011. Multi-source transfer of delexicalized dependency parsers. In *Proceedings of the Conference on Empirical Methods in Natural Language Processing*, EMNLP '11, page 62–72, USA. Association for Computational Linguistics.

Nguyen, Toan Q. and David Chiang. 2017. Transfer learning across low-resource, related languages for neural machine translation. In *International Joint Conference on Natural Language Processing*.

Nooralahzadeh, Farhad, Giannis Bekoulis, Johannes Bjerva, and Isabelle Augenstein. 2020. Zero-shot cross-lingual transfer with meta learning. In *Proceedings of the 2020 Conference on Empirical Methods in Natural Language Processing (EMNLP)*, pages 4547–4562, Online, November. Association for Computational Linguistics.

Papineni, Kishore, Salim Roukos, Todd Ward, and Wei-Jing Zhu. 2002. Bleu: a method for automatic evaluation of machine translation. In *Proceedings of the 40th Annual Meeting of the Association for Computational Linguistics*, pages 311–318, Philadelphia, Pennsylvania, USA, July. Association for Computational Linguistics.

Tang, Yuqing, Chau Tran, Xian Li, Peng-Jen Chen, Naman Goyal, Vishrav Chaudhary, Jiatao Gu, and Angela Fan. 2020. Multilingual translation with extensible multilingual pretraining and finetuning.

Wolf, Thomas, Lysandre Debut, Victor Sanh, Julien Chaumond, Clement Delangue, Anthony Moi, Pierric Cistac, Tim Rault, Remi Louf, Morgan Funtowicz, Joe Davison, Sam Shleifer, Patrick von Platen, Clara Ma, Yacine Jernite, Julien Plu, Canwen Xu, Teven Le Scao, Sylvain Gugger, Mariama Drame, Quentin Lhoest, and Alexander Rush. 2020. Transformers: State-of-the-art natural language processing. In *Proceedings of the 2020 Conference on Empirical Methods in Natural Language Processing: System Demonstrations*, pages 38–45, Online, October. Association for Computational Linguistics.

Zhang, Biao, Philip Williams, Ivan Titov, and Rico Sennrich. 2020. Improving massively multilingual neural machine translation and zero-shot translation. In *Proceedings of the 58th Annual Meeting of the Association for Computational Linguistics*, pages 1628–1639, Online, July. Association for Computational Linguistics.

Zoph, Barret, Deniz Yuret, Jonathan May, and Kevin Knight. 2016. Transfer learning for low-resource neural machine translation. In *Conference on Empirical Methods in Natural Language Processing*.